\documentclass{article}

\usepackage{spconf,amsmath,graphicx,hyperref} 

\usepackage{placeins}     
\usepackage{indentfirst}  
\usepackage{svg}          
\usepackage{float}        
\usepackage{caption}      
\usepackage{ragged2e}     

\usepackage{amssymb}      

\usepackage{booktabs}     
\usepackage{hhline}       
\usepackage{makecell}     
\usepackage{tabularx}     
\usepackage{arydshln}     

\usepackage{cleveref}     

\newcolumntype{C}{>{\Centering\arraybackslash}X}
\newcolumntype{L}{>{\raggedright\arraybackslash}X}

\newcolumntype{C}{>{\Centering\arraybackslash}X}
\newcolumntype{L}{>{\raggedright\arraybackslash}X}
\title{PAGS: Priority-Adaptive Gaussian Splatting for Dynamic Driving Scenes}
\name{Ying A$^{1,3}$, Wenzhang Sun$^{2}$, Chang Zeng$^{2}$, Chunfeng Wang$^{2}$, Hao Li$^{2}$, Jianxun Cui$^{1,3}$}

\address{$^{1}$ Harbin Institute of Technology, China\\
$^{2}$ Li Auto, China\\
 $^{3}$Chongqing Research Institute of HIT, China}
\begin{document}
\ninept
\maketitle
\begin{abstract}
Reconstructing dynamic 3D urban scenes is crucial for autonomous driving, yet current methods face a stark trade-off between fidelity and computational cost. This inefficiency stems from their semantically agnostic design, which allocates resources uniformly, treating static backgrounds and safety-critical objects with equal importance. To address this, we introduce Priority-Adaptive Gaussian Splatting (PAGS), a framework that injects task-aware semantic priorities directly into the 3D reconstruction and rendering pipeline. PAGS introduces two core contributions: (1) Semantically-Guided Pruning and Regularization strategy, which employs a hybrid importance metric to aggressively simplify non-critical scene elements while preserving fine-grained details on objects vital for navigation. (2) Priority-Driven Rendering pipeline, which employs a priority-based depth pre-pass to aggressively cull occluded primitives and accelerate the final shading computations. Extensive experiments on the Waymo and KITTI datasets demonstrate that PAGS achieves exceptional reconstruction quality, particularly on safety-critical objects, while significantly reducing training time and boosting rendering speeds to over 350 FPS.
\end{abstract}
\begin{keywords}
Gaussian Splatting, Real-Time Rendering, Autonomous Driving
\end{keywords}
\section{Introduction}
\label{sec:intro}

The reconstruction of dynamic, large-scale urban environments is a cornerstone of modern autonomous driving systems, providing the foundation for critical applications like simulation testing, synthetic data generation, and the creation of digital twins \cite{yang2023unisim,wu2023mars}. With the advent of 3D Gaussian Splatting \cite{kerbl20233d}, the field has seen a significant leap forward, enabling real-time, photorealistic synthesis of novel views. To contend with the complexity of bustling cityscapes, a dominant paradigm has emerged: decomposing the scene into a static background and multiple, independently modeled dynamic foregrounds \cite{ost2021neural,kundu2022panoptic,tancik2022block,wu20244d}. This strategy has been adeptly adopted by contemporary 3DGS-based methods such as StreetGS \cite{yan2024street} and DrivingGaussian \cite{zhou2024drivinggaussian}.

\begin{figure}[ht]
\centering
\includegraphics[width=0.5\textwidth]{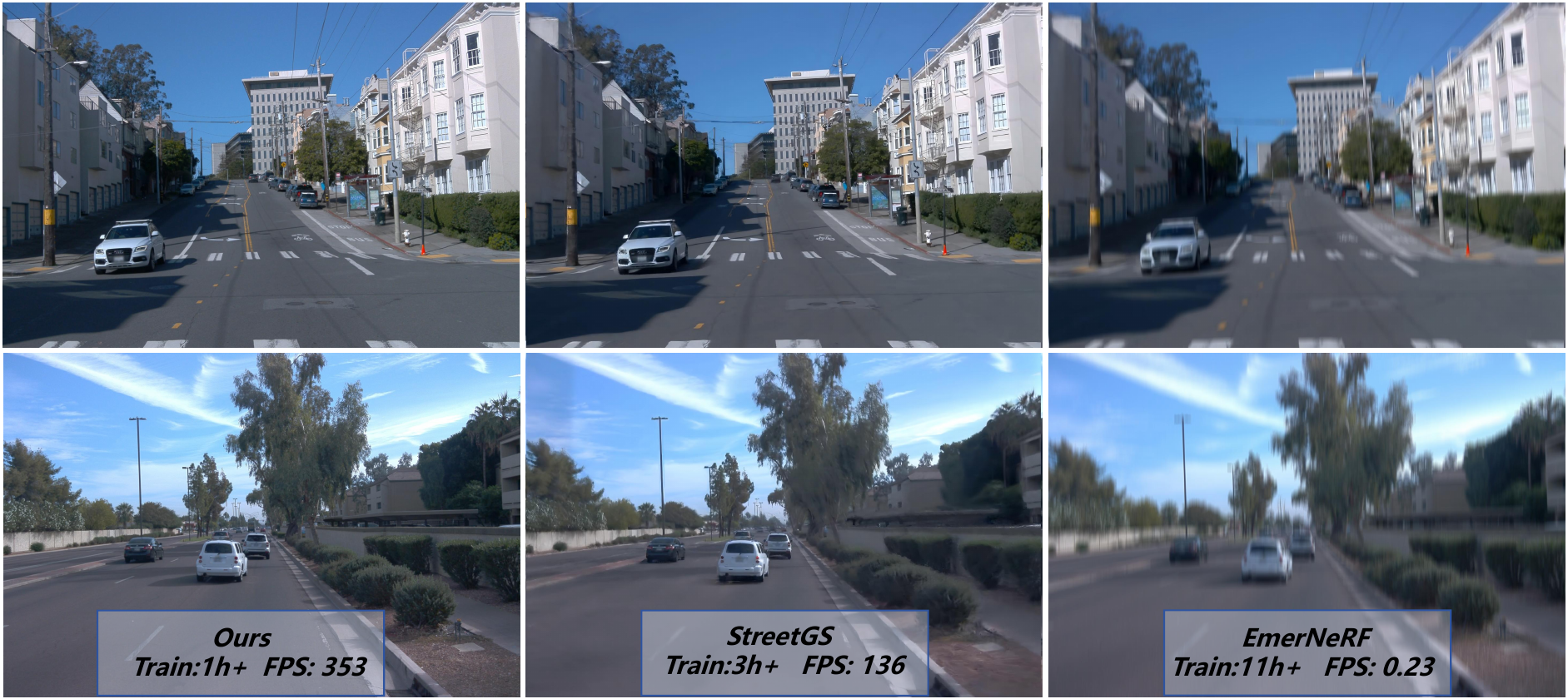}
\caption{Qualitative comparisons on the Waymo \cite{sun2020scalability} dataset. Our method achieves a sharper and more detailed reconstruction with just over \textbf{1 hour} of training. This quality surpasses both StreetGS \cite{yan2024street} and EmerNeRF \cite{yang2023emernerf}, which produce less detailed results despite requiring significantly longer training times of over \textbf{3h} and \textbf{11h}.}
\label{fig:fistduibi}
\vspace{-3mm}
\end{figure}

However, the pursuit of universal high fidelity reveals a fundamental conflict when applied to autonomous driving. While recent works such as Speedy-Splat \cite{hanson2025speedy}, FlashGS \cite{feng2025flashgs}, and Mini-Splatting \cite{fang2024mini} have made significant strides in acceleration, they operate under a paradigm of uniform optimization. These methods remain semantically agnostic, failing to distinguish between functionally critical and non-critical scene components. This semantic blindness leads to a profound misallocation of resources.  Significant computational budget is expended on perfecting the texture of a distant, static building or the foliage of a roadside tree—elements that have negligible impact on driving decisions. Every computation cycle spent refining a non-critical background element is a cycle not spent capturing the high-frequency details of a safety-critical foreground object. The finite representational capacity, when stretched thin across an entire scene, inevitably leads to compromised fidelity where it matters most: on pedestrians, cyclists, and other vehicles \cite{liu2025hi}. Consequently, subtle but vital visual cues risk being smoothed over or lost in a blur of generalized detail.

\begin{figure*}[htb]
\centering
\includegraphics[width=1\textwidth]{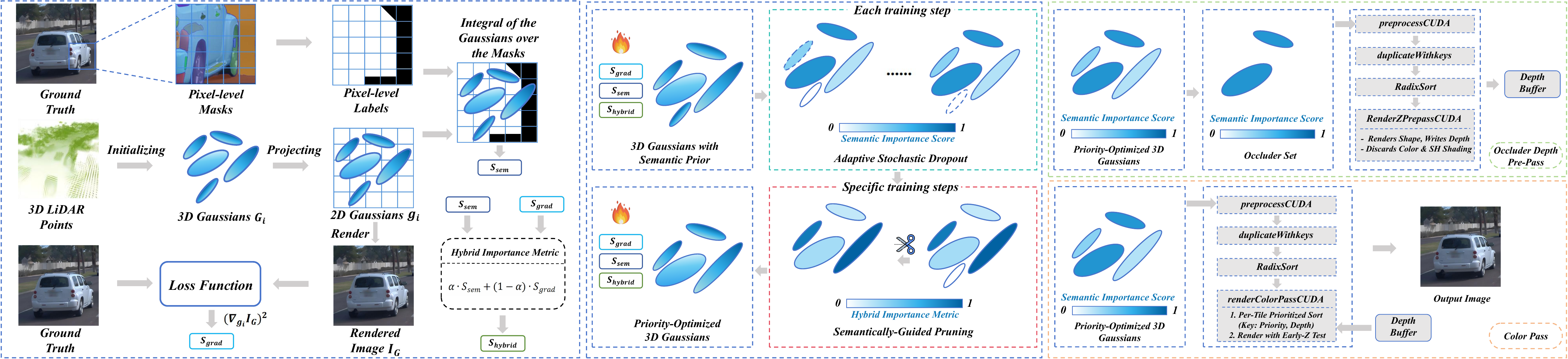}

\caption{\textbf{Overview of our proposed framework.} Our pipeline embeds semantic importance throughout the reconstruction and rendering process. \textbf{(Left)} We begin with an offline semantic scene decomposition, using foundation models to assign a semantic prior to each 3D Gaussian. \textbf{(Center)} During training, this semantic prior guides two synergistic optimization strategies: Adaptive Stochastic Dropout, which applies regularization based on semantic importance scores in each iteration, and Semantically-Guided Pruning, which uses a hybrid importance metric to strategically remove Gaussians at specific intervals. This yields a set of priority-optimized 3D Gaussians. \textbf{(Right)} At inference, a priority-driven rendering pipeline first generates a depth map using high-importance occluders, then leverages it to accelerate the final color pass through efficient hardware-based culling.}
\vspace{-5mm}
\label{fig:pipeline_overview}
\end{figure*}

To bridge this critical gap, we introduce Priority-Adaptive Gaussian Splatting (PAGS), a framework that embeds task-oriented semantic importance throughout the reconstruction and rendering pipeline. PAGS materializes this concept through two synergistic contributions. First, \textbf{Semantically-Guided Pruning and Regularization} strategy employs a hybrid metric—balancing a static, foundation model-derived semantic score with a dynamic, gradient-based contribution. This allows for aggressive simplification of non-critical backgrounds while preserving high-fidelity details on key objects like vehicles and pedestrians. Second, the \textbf{Priority-Driven Rendering} pipeline first uses high-importance primitives to generate a coarse depth map. The GPU's hardware-accelerated Early-Z test then leverages this map to cull occluded fragments before they undergo expensive shading. This significantly boosts rendering speed without compromising the perceptual quality of critical scene elements. Extensive experiments on the Waymo and KITTI \cite{geiger2012we} datasets demonstrate the efficacy of our approach. PAGS achieves a superior balance between reconstruction fidelity and computational efficiency, outperforming existing methods. Specifically, our framework not only enhances the visual quality of safety-critical objects but also drastically reduces training time while achieving real-time rendering speeds exceeding 350 FPS. 


\section{Methods}
\label{sec:format}

\subsection{Compositional gaussian representation}
We decompose the scene into distinct static and dynamic components. The static background is represented by a set of 3D Gaussian fixed in the world coordinate, initialized from a combination of LiDAR scans and Structure-from-Motion (SfM) \cite{schonberger2016structure} point clouds to ensure comprehensive coverage. Each primitive is defined by its position ($\mu_{b}$), covariance ($\Sigma_{b}$), opacity ($\alpha_{b}$), and view-dependent appearance modeled by Spherical Harmonics (SH). For unbounded regions like the sky, we employ a high-resolution cubemap \cite{tancik2022block}. 

Concurrently, each moving object is modeled as an independent set of Gaussian primitives within its own local coordinate frame. These primitives share similar attributes to their static counterparts, such as opacity ($\alpha_{o}$) and local scale ($S_{o}$). To capture motion, we associate each object with a series of optimizable, time-varying poses, each comprising a rotation $R_{c}$ and translation $T_{c}$. To realistically model the time-varying appearance of dynamic objects under changing illumination, we utilize a four-dimensional Spherical Harmonics model, where each SH coefficient is represented as a function of time through a set of Fourier coefficients \cite{yan2024street}.

\subsection{Semantically-guided pruning and regularization}

\subsubsection{Offline semantic scene decomposition}

The foundation of our priority-adaptive framework is a clear semantic understanding of the scene. As a one-time preprocessing step, we first employ the Segment Anything Model (SAM) \cite{kirillov2023segment} to generate class-agnostic instance masks for all images. To assign semantic labels to these masks, we integrate results from an off-the-shelf semantic segmentation model \cite{radford2021learning}. We group these semantic labels into a principled binary partition: categories such as vehicles, pedestrians, and cyclists are designated as 
Critical, while all others (e.g., buildings, roads, vegetation) are classified as Non-Critical.

\begin{table*}[t]
\centering
\caption{Quantitative results on the Waymo and KITTI datasets.}
\vspace{-6pt}
\label{tab:performance_comparison}
\resizebox{\textwidth}{!}{%
\begin{tabular}{l ccccc ccccc}
\toprule
\textbf{Dataset}    & \multicolumn{5}{c}{\textbf{Waymo Open Dataset}} & \multicolumn{5}{c}{\textbf{KITTI Dataset}} \\
\cmidrule(lr){2-6} \cmidrule(lr){7-11}
\textbf{Method} & \textbf{PSNR} $\uparrow$ & \textbf{SSIM} $\uparrow$ & \textbf{LPIPS} $\downarrow$ & \textbf{FPS} $\uparrow$ & \textbf{Train Time} $\downarrow$ & \textbf{PSNR} $\uparrow$ & \textbf{SSIM} $\uparrow$ & \textbf{LPIPS} $\downarrow$ & \textbf{FPS} $\uparrow$ & \textbf{Train Time} $\downarrow$ \\
\midrule
EmerNeRF\cite{yang2023emernerf}    & 28.11 & 0.786 & 0.377 & 0.23 & 11h48m & 26.95 & 0.828 & 0.219 & 0.28 & 11h51m \\
PVG\cite{chen2023periodic}  & 30.46 & 0.910 & 0.229 & 52 & 4h23m & 31.82 & 0.937 & 0.070 & 59 & 4h37m \\
StreetGS\cite{yan2024street}    & 32.21 & 0.907 & \textbf{0.073} & 136 & 3h39m & 32.76 & 0.922 & 0.067 & 141 & 3h54m \\
DeSiRe-GS\cite{peng2025desire}  & 33.52 & 0.917 & 0.204 & 36 & 5h34m & 33.75 & 0.934 & 0.044 & 41 & 5h47m \\
\textbf{Ours}       & \textbf{34.63} & \textbf{0.933} & \textbf{0.073} & \textbf{353} & \textbf{1h22m} & \textbf{34.58} & \textbf{0.947} & \textbf{0.032} & \textbf{365} & \textbf{1h31m} \\
\bottomrule
\end{tabular}%
}
\end{table*}

\begin{figure*}[t]
\centering
\vspace{-6pt}
\includegraphics[width=\textwidth]{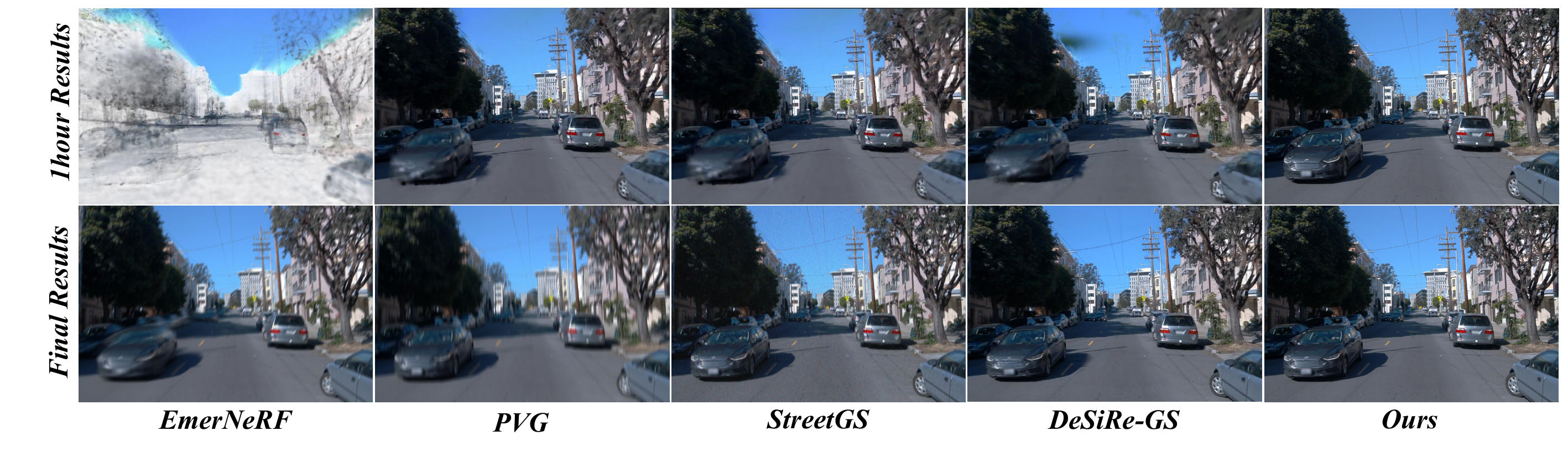}
\caption{\textbf{Qualitative results on the Waymo dataset}. The \textbf{bottom row} shows the final, fully converged reconstruction results for all methods. Our approach produces significantly sharper details on both dynamic objects and the static background. The \textbf{top row} presents a time-equalized comparison, showing the output of each method after an identical, fixed training duration. This highlights the superior convergence speed of our method, which achieves a high-fidelity result while competitors still exhibit noticeable blur and artifacts.}
\vspace{-3mm}
\label{fig:full_comparison}
\end{figure*}

\begin{table}[ht]
\begin{minipage}{0.5\textwidth}
\centering

\caption{Inference efficiency and memory footprint comparison.}
\vspace{-6pt}
\label{tab:inference_efficiency_comparison}
\small
\setlength{\tabcolsep}{4pt}
\begin{tabularx}{\linewidth}{L C C C}
\toprule
Method & Speed (FPS) $\uparrow$ & Size (MB) $\downarrow$ & VRAM (GB) $\downarrow$ \\
\midrule
EmerNeRF & 0.23 & 1217 & 10.5 \\
PVG & 52 & 959 & 8.2 \\
StreetGS  & 136 & 853 & 7.8 \\
DeSiRe-GS & 36 & 984 & 8.5 \\
\textbf{Ours} & \textbf{353} & \textbf{530} & \textbf{6.1} \\
\bottomrule
\end{tabularx}
\end{minipage}
\vspace{-5mm}
\end{table}

\subsubsection{Semantically-guided pruning}

To intelligently allocate model capacity, our framework employs a semantically-guided pruning strategy based on a Hybrid Importance Metric ($S_{\text{hybrid}}$). This metric balances a stable, top-down semantic prior with a dynamic, data-driven contribution score, ensuring that pruning decisions are aligned with task priorities.

The metric consists of two components. The first is a Semantic Importance Score ($S_{\text{sem}}$), calculated once during initialization. For each gaussian primitive, we project it into all views and compute the average overlap with pre-defined semantic masks. This yields a score $S_{\mathrm{sem}}\in[0,1]$ that acts as a strong semantic prior throughout training \cite{siddiqui2023panoptic}. The second component is a Dynamic Contribution Score ($S_{\text{grad}}$), which quantifies a primitive's instantaneous importance to the reconstruction. This score is derived by aggregating the squared gradients of the final pixel color with respect to the primitive's contribution, which are readily available during the backward pass. The two scores are combined via a hyperparameter $\alpha$:
\begin{equation}
\label{eq:contribution_score}
S_{\text{hybrid}} = \alpha \cdot S_{\text{sem}} + (1 - \alpha) \cdot S_{\text{grad}}
\end{equation}
This formulation allows us to apply a more lenient pruning threshold to primitives with high semantic importance, preserving fine details on critical objects even when their immediate gradient contribution is low, while non-critical elements are pruned more aggressively.

Operationally, we use $S_{\text{hybrid}}$ to rank and prune primitives at specific training intervals. During the densification stage (at 10k, 15k, and 20k iterations), we prune 60\% of the Gaussians with the lowest scores. Subsequently, during fine-tuning (starting at 25k iterations), we continue to prune 30\% of primitives every 5k iterations. As demonstrated in our ablation study (\Cref{sec:Pruning rate sensitivity analysis}), these specific rates are critical for achieving an optimal balance between training efficiency and final reconstruction quality.

\subsubsection{Adaptive stochastic dropout}

Reconstructing moving objects is challenging due to sparse views, which risks overfitting and degrading reconstruction quality. To mitigate this, we introduce Adaptive Stochastic Dropout, a regularization technique that respects the scene's semantic hierarchy. The dropout probability for each primitive is inversely modulated by its semantic importance, as defined by the formula:
\vspace{-2mm}
\begin{equation}
\label{eq:contribution_score}
D_{t,i} = \left(1 - \beta \cdot S_{\mathrm{sem},i}\right) \cdot \gamma \cdot \frac{t}{t_{\mathrm{total}}}
\vspace{-2mm}
\end{equation}
This approach applies gentler regularization to critical, sparsely-observed objects, preventing overfitting while preserving their fine details. This method works synergistically with our pruning strategy, as primitives that are frequently dropped become natural candidates for removal. To maintain color fidelity, the opacities of surviving primitives are amplified by a compensation factor:
\vspace{-2mm}
\begin{equation}
\label{eq:contribution_score}
C(i) = \frac{1}{1-D_{t,i}}
\vspace{-2mm}
\end{equation}
\subsection{Priority-driven rendering}\label{sec:rendering_pipeline}
To translate semantic importance into rendering efficiency, we introduce a Priority-Driven Rendering pipeline that leverages hardware occlusion culling. The first pass, an Occluder Depth Pre-Pass, rapidly establishes a coarse depth map of the scene. This is achieved by rendering a filtered subset of primitives, termed the Occluder Set, which are selected based on high semantic scores ($S_{\text{sem}>}$0.5) and opacity. This pass utilizes a minimal shader that discards all color and view-dependent calculations, writing only depth values to the Z-buffer with negligible computational overhead.

The subsequent Color Pass renders the final, high-fidelity image for all primitives, capitalizing on the depth information from the pre-pass to achieve significant acceleration. Specifically, the GPU's hardware-accelerated Early-Z test compares the depth of each incoming fragment against the pre-populated Z-buffer, instantly culling any occluded fragments before expensive shading computations. To ensure correct alpha blending for the semi-transparent Gaussian model, a per-tile, back-to-front sort is performed using a composite key derived from both semantic priority and depth. By eliminating the redundant workload of rendering occluded primitives, this pipeline focuses computation exclusively on visible surfaces, which is the primary driver for the substantial increase in rendering speed to over 350 FPS in our results.

\section{Experiments}
\label{sec:pagestyle}

\subsection{Experimental setup}
We conduct evaluations on the Waymo and KITTI benchmarks. For Waymo, we use three frontal cameras, while for KITTI, we use the stereo camera pair. We evaluate reconstruction fidelity and perceptual quality using PSNR, SSIM, and LPIPS \cite{zhang2018unreasonable}, and assess efficiency via total training time and rendering speed (FPS). Our method is compared against advanced approaches including EmerNeRF \cite{yang2023emernerf}, PVG \cite{chen2023periodic}, StreetGS \cite{yan2024street}, and DeSiRe-GS \cite{peng2025desire}. All experiments are performed on a NVIDIA RTX 4090 GPU. For our framework, key hyperparameters are set to $\alpha$=0.4 for the hybrid importance metric, $\gamma$=0.25 and $\beta$=0.5 for adaptive stochastic dropout.

\subsection{Quantitative and qualitative analysis}
Quantitatively, as shown in \Cref{tab:performance_comparison}, our method achieves a PSNR of 34.63 and an SSIM of 0.933 on the Waymo dataset, surpassing all compared methods while significantly accelerating rendering speed to 353 FPS. Notably, our training time of 1h22m is significantly shorter than most competing approaches. Similar performance gains are observed on the KITTI dataset. In terms of inference efficiency and memory footprint, as detailed in \Cref{tab:inference_efficiency_comparison}, our framework demonstrates a significant advantage crucial for practical deployment. PAGS achieves a real-time rendering speed of 353 FPS while maintaining a compact model size of just 530 MB and a low VRAM footprint of 6.1 GB, significantly outperforming all competing methods. This superior performance is a direct outcome of our core designs; the Semantically-Guided Pruning strategy is instrumental in creating a highly compact model, while the Priority-Driven Rendering pipeline is the primary driver behind the substantial leap in rendering speed.

Qualitatively, as shown in \Cref{fig:full_comparison}, the visual comparison further underscores our method's superiority. In the final results, our approach produces reconstructions with significantly sharper and clearer details on both vehicles and background elements. After just one hour of training, our method yields a high-fidelity result that is already much clearer than the blurry and artifact-laden outputs from competing methods at the same training duration. This combined evidence demonstrates the effectiveness and efficiency of our proposed framework.

\begin{figure}[t]
\centering
\includegraphics[width=\linewidth]{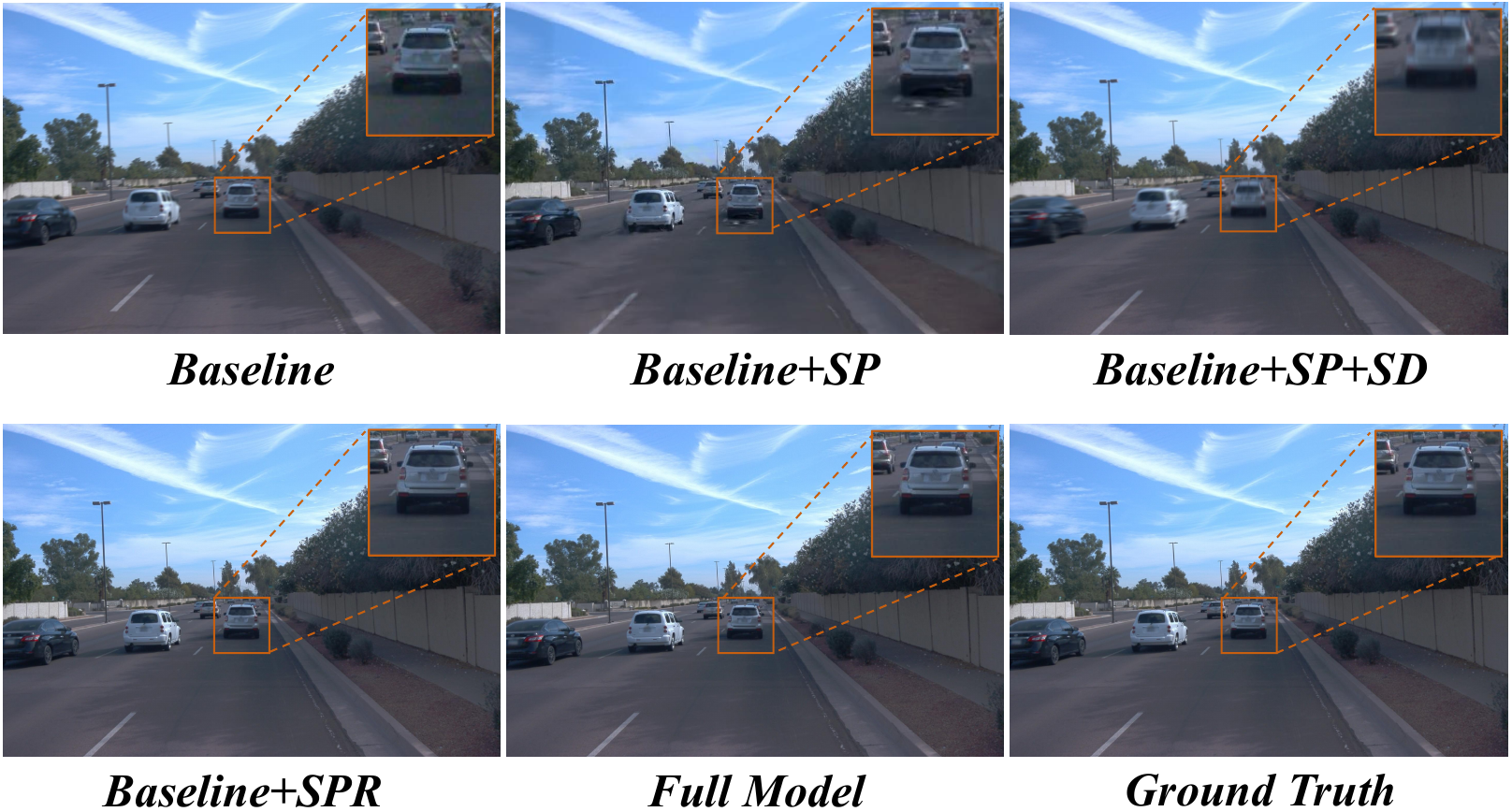}
\caption{\textbf{Qualitative ablation of key components on the Waymo dataset.} Our Semantically-Guided Pruning and Regularization (SPR) strategy yields a markedly sharper reconstruction of the vehicle than the baseline combining Staged Pruning (SP) and Stochastic Dropout (SD).}
\label{fig:ablation_qualitative_duibitu}
\vspace{-3mm}
\end{figure}

\begin{table}[t]
\hfill
\begin{minipage}{0.5\textwidth}
    \centering
     
    \caption{Quantitative ablation of key components.}
    \vspace{-6pt}
    \label{tab:ablation_components}
    
    \small
    \setlength{\tabcolsep}{3pt}
    
    \begin{tabular}{cccc|ccccc}
    
    \hline
    
    \textbf{SP} & \textbf{SD} & \textbf{SPR} & \textbf{PDR} & \textbf{PSNR} & \textbf{SSIM} & \textbf{LPIPS} & \textbf{FPS} & \textbf{Time} \\
    
    \hline
    
    \checkmark & & & & 29.83 & 0.892 & 0.083 & 135 & 1h30m \\
    & \checkmark & & & 32.87 & 0.951 & 0.041 & 130 & 3h35m \\
    \checkmark & \checkmark & & & 32.35 & 0.953 & 0.039 & 134 & 1h36m \\
    & & \checkmark & & \textbf{34.95} & \textbf{0.958} & 0.035 & 134 & 1h24m \\
    & & & \checkmark & 32.02 & 0.942 & \textbf{0.031} & 324 & 3h20m \\
    & & \checkmark & \checkmark & 34.63 & 0.933 & 0.073 & \textbf{353} & \textbf{1h22m} \\
    \hline
    \end{tabular}
\end{minipage}
\vspace{-2mm}
\end{table}

\subsection{Ablation studies}\label{sec:ablation_studies}
\subsubsection{Analysis of component contributions}
We validates the distinct contributions of our Semantically-Guided Pruning and Regularization (SPR) and Priority-Driven Rendering (PDR) modules. We first benchmarked our SPR module against a non-semantic baseline that combines standard Staged Pruning (SP) and Stochastic Dropout (SD). As detailed in \Cref{tab:ablation_components}, the baseline achieves a PSNR of 32.35 in 1h36m of training. In contrast, our SPR module elevates the PSNR to 34.95 while reducing training time to just 1h24m. This quantitative superiority is further supported by our qualitative results in \Cref{fig:ablation_qualitative_duibitu}, which shows that the SPR module yields a markedly sharper reconstruction of critical objects compared to the baseline. This demonstrates that embedding semantic importance breaks the conventional trade-off, yielding a model that is both significantly more accurate and faster to train.

Furthermore, our Priority-Driven Rendering pipeline is designed for inference acceleration. 
\begin{figure}[t]
\centering
\includegraphics[width=0.5\textwidth]{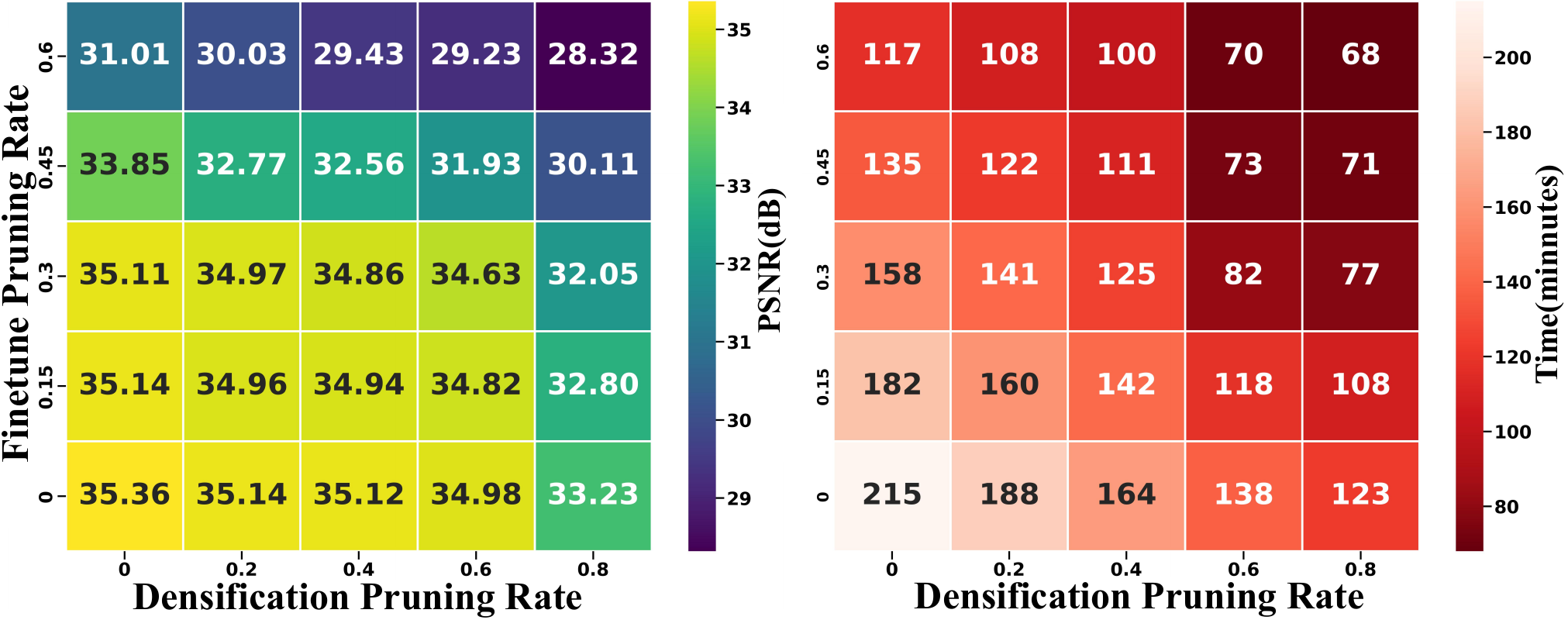}
\caption{Impact of pruning rates on PSNR and training time.}
\label{fig:relitu}
\vspace{-3mm}
\end{figure}
By applying PDR to the SPR-trained model, rendering speed is boosted by 2.6x from 134 FPS to 353 FPS. This real-time performance is enabled by a controlled trade-off, which results in a final global PSNR of 34.63. Consequently, the complete PAGS framework provides a highly optimized solution, achieving a superior balance of reconstruction quality, training efficiency, and rendering speed.

\subsubsection{Analysis of pruning rate sensitivity}\label{sec:Pruning rate sensitivity analysis}
For Semantically-Guided Pruning, We conducted a sensitivity analysis to determine the optimal configuration for our key hyperparameters. We investigated the impact of densification and fine-tuning pruning rates on quality and training time. As visualized in \Cref{fig:relitu}, we identified rates of 0.6 for densification and 0.3 for fine-tuning as the optimal balance, achieving a competitive PSNR of 34.63 while reducing training time to 82 minutes. 



\begin{table}[t]
\begin{minipage}{0.5\textwidth}
    \centering
    
    \caption{Validation of the hybrid importance metric.}
    \vspace{-6pt}
    \label{tab:hybrid_metric_ablation_half_tabularx}
    \small
    \setlength{\tabcolsep}{4pt}

    \begin{tabularx}{\linewidth}{L c C C C C}
    \toprule

    Methods & $\alpha$ & PSNR-C  & PSNR-NC & PSNR  & Gauss.(M)  \\
    \midrule

    Gradient & 0.0 & 28.15 & \textbf{34.54} & 33.38 & 58.8 \\
    Semantic & 1.0 & 30.98 & 27.82 & 29.15 & \textbf{48.9} \\
    \textbf{Hybrid} & 0.4 & \textbf{35.97} & 33.20 & \textbf{34.63} & 52.4\\
    \bottomrule
    \end{tabularx}
\end{minipage}
\vspace{-5mm}
\end{table}

\subsubsection{Analysis of hybrid importance metric}

To validate our core innovation—the hybrid importance metric —we compare it against two specialized variants: a Gradient-Only strategy and a Semantic-Only strategy. The results in \Cref{tab:hybrid_metric_ablation_half_tabularx} validate our approach. The Gradient-Only approach neglects the fidelity of smaller, critical elements, resulting in a low PSNR-Critical score. Conversely, the Semantic-Only variant preserves critical objects at the expense of background integrity, leading to the lowest overall PSNR. Our hybrid approach   emerges as the optimal solution, achieving the highest overall PSNR (34.63) and the best PSNR-Critical (35.97) while producing a compact model (52.4M Gaussians).

\section{Conclusion}
We presented Priority-Adaptive Gaussian Splatting (PAGS), a framework designed to overcome the semantic-agnostic limitations of current 3DGS methods for dynamic driving scenes. By synergizing a semantically-guided pruning and and Regularization strategy with a priority-driven rendering pipeline, PAGS effectively concentrates computation on safety-critical elements. Our results on the Waymo and KITTI datasets confirm the benefits of this approach, achieving rendering speeds over 350 FPS and reduced training times without compromising reconstruction quality on key objects. PAGS demonstrates that task-aware optimization is a key enabler for deploying high-fidelity, real-time 3D Gaussian Splatting in practical autonomous driving applications.

\label{sec:typestyle}

\bibliographystyle{IEEEbib}

\end{document}